\documentclass{bmvc2k}
\usepackage[utf8]{inputenc}
\usepackage{float}
\usepackage{multirow}

\title{Predicting Visual Memory Schemas with Variational Autoencoders}

\addauthor{Cameron Kyle-Davidson*}{ckd505@york.ac.uk}{1}
\addauthor{Adrian Bors*}{adrian.bors@york.ac.uk}{1}
\addauthor{Karla Evans**}{karla.evans@york.ac.uk}{2}

\addinstitution{
 *Dept. of Computer Science\\
 *YO10 5GH \\
 **Dept. of Psychology\\
 **YO10 5DD\\
 University of York\\
 York, UK\\
}

\runninghead{Kyle-Davidson, Bors, Evans}{Predicting VMS with Variational Autoencoders}

\def\etal{\emph{et al}\bmvaOneDot}

\begin{document}

\maketitle

\begin{abstract}
Visual memory schema (VMS) maps show which regions of an image cause that image to be remembered or falsely remembered. Previous work has succeeded in generating low resolution VMS maps using convolutional neural networks. We instead approach this problem as an image-to-image translation task making use of a variational autoencoder. This approach allows us to generate higher resolution dual channel images that represent visual memory schemas, allowing us to evaluate predicted true memorability and false memorability separately. We also evaluate the relationship between VMS maps, predicted VMS maps, ground truth memorability scores, and predicted memorability scores.
\end{abstract}

\section{Introduction}
\label{sec:intro}
Determining capacity and the nature of visual memory has been a focus of psychological experiments for decades. However, it is only recently that \textit{memorability} has been able to be defined and predicted using computational methods. This definition of memorability has been found to be separate to other commonly computed image factors such as saliency or interestingness. The basis of this definition in prior work is related to the hit rate (HR) of an image, which is how well a target image is recognised after being repeated in a sequence of images.
Predicting the memorability score for an image representing how likely a given image is to be remembered by a human during a recognition test, is a difficult task - memorability has been shown to be associated with the semantic content of the image, a complex dimension to extract. With the advent of large memorability datasets that contain tens of thousands of images paired with ground truth memorability scores, recent deep learning models' have succeeded in achieving close-to-human performance in predicting how likely an image is to be remembered.

\begin{figure}[H]
\vspace*{-0.2cm}
\centering
\includegraphics[width=0.75\textwidth]{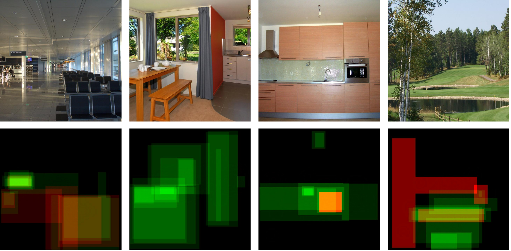}
\caption{Examples of images and their corresponding VMS maps. In the second row of images, red areas correspond to regions that cause the associated image to be falsely remembered, while green regions are responsible for correctly remembering the image. Falsely remembered regions cause a person to believe they have seen the given image when in fact, they have not.}
\label{Fig1}
\vspace*{-0.3cm}
\end{figure}

Previous work in the arena of memorability prediction has been engineered with the goal of predicting memorability scores for a given image. Few research studies explored creating models capable of predicting the regions of an image that contribute the most to an image's memorability. These models' predictions of memorable regions lack a clear relation to the ground truth, as until very recently no dataset of the regions that cause \textit{humans} to find a given image memorable, existed. A new image memorisation dataset was introduced in \cite{akagunduz_defining_2019} which tackles this problem by introducing the \textit{VISCHEMA} image dataset, which contains human indications of what regions made them remember certain images. Moreover, a new concept known as Visual Memory Schema (VMS), which associates for each image in the dataset two dimensional probability density functions (PDF) that represent which areas of that image cause to be either remembered (a true VMS map), or falsely remembered (a false VMS map). Examples of the VMSs are shown on the second row in Fig.~\ref{Fig1}, corresponding to the images from above. According to the experiments, such VMS maps have been shown to be consistent between people. By following certain psychology studies it was hypothesised in \cite{akagunduz_defining_2019} that these regions correspond to mental structures that aid remembering.

We hypothesise that a relationship exists between images that have strong, either true or false VMS maps, and seek to learn this relationship to better understand the meaning of the VMS maps and the relation between memorability and false memorability. More specifically, we expect that memorable images align along dimensions of `memorability' and likewise for false memorability. Learning a structured embedding in this `memorability space' would lead to the capability to generate both true and false VMS maps for unseen images, and hence aid understanding of which mental structures contribute to remembering or falsely remembering an image. We approach this problem via Variational Autoencoders (VAE) models, which have previously been shown to be capable of learning to group similar data in an unsupervised fashion by mapping through a latent space. We hence pose the problem of learning such transformations, and the resulting VMS map generation, as an image-to-image translation problem.

A VAE model, based upon human ground truth data, that determines an image to be remembered or falsely remembered, is proposed in this paper.
We examine this model and verify that features that lead to both strong true VMS maps and strong false VMS maps tend to be grouped together, respectively. Our experiments explore the produced VMS maps of this model over a new dataset with identical categories to the original VISCHEMA dataset. We conclude that saliency alone is not the driving feature for this approach. The structure of the paper is as follows:  Section~\ref{Sec2} outlines the previous work in predicting memorability for images, Section~\ref{VAE} presents the Variational Autoencoder,
Section~\ref{Sec4} describes how we use VAE for learning VMS maps, Section~\ref{Exp} presents the experimental results and Section~\ref{Con} draws the conclusions of this study.

\vspace*{-0.2cm}
\section{Relevant Work}
\label{Sec2}
\vspace*{-0.2cm}
\subsection{Predicting the memorability of images}
\vspace*{-0.1cm}

Predicting how likely an image is to be remembered is a problem that has only recently become an active area of interest in computer vision. Isola \etal{} created an initial memorability dataset of over 2000 images and experimented with using certain feature extractors paired with a support vector machine (SVM) for prediction\cite{isola_what_2011}. Isola found that humans generally agree on what is memorable, achieving a consistency of more than 0.68 as measured by the Spearman Rank Correlation metric. In general hand picked features achieve a consistency of less than 0.5. Peng \etal{} \cite{peng_predicting_2015} and Jing \etal{} \cite{jing_predicting_2017} use multiview modelling achieving a consistency with the ground truth greater than that of any SVM based model. Later work by Khosla \etal{} improves upon these results, introducing the LaMem dataset \cite{khosla_understanding_2015} of 60,000 images and their corresponding memorability scores. Moreover, they introduced MemNet, a convolutional neural network (CNN), for the purpose of prediction. Fajtl \etal{} constructed a CNN-LSTM (Long Short-Term Memory) hybrid model known as AMNet that iteratively generates attention-based memorability scores, achieving a performance very close to human consistency \cite{fajtl_amnet:_2018} when trained upon the LaMem dataset.

\vspace*{-0.2cm}
\subsection{Predicting memorability maps}
\vspace*{-0.1cm}

Relatively little work has examined the generation of memorability maps directly. Khosla \etal{} used a probabilistic process to generate memorability maps \cite{khosla_memorability_2012} by considering the regions of images that are likely to be remembered or forgotten. The MemNet CNN developed also by Khosla was also capable of creating heatmaps of the most memorable and the least memorable regions of a given image. Similarly, the work of Fajtl \etal{} iteratively generates attention based memory maps that are concatenated to generate a final score. However, none of these methods would generate memory maps which can be compared with ground truth maps of memorability.

A dataset of 800 scene images and their associated `Visual Memory Schema' (VMS) was developed during the VISCHEMA experiment. The images considered for this dataset are divided through a tree structure, where each level describes a certain aspect of that image in increasing detail. Images are first divided into \textit{indoor} and \textit{outdoor}. The \textit{indoor} category contains the categories \textit{private} and \textit{public} while the \textit{outdoor} category contains \textit{man-made} and \textit{natural} images. These are further subdivided, with \textit{private} containing \textit{kitchen} and \textit{living room}, \textit{public} containing \textit{small} and \textit{big} (which refers to the size of the public space shown in the image). The \textit{man-made} category contains \textit{work/home} and \textit{public entertainment} and the \textit{natural} category divides images into being either \textit{populated} or \textit{isolated} regions. A Visual Memory Schema defines the regions of an image that led to that image being either remembered or falsely remembered. These VMS maps represent the cognitive elements shared by people that influence the memorability of a given image. True VMSs have a high degree of consistency while False VMSs have a lower degree of consistency. In the research study from \cite{akagunduz_defining_2019} a pretrained VGG network is fine tuned to reconstruct VMS maps at a $20 \times 20$ resolution. However, the results from \cite{akagunduz_defining_2019} do not predict separately true or false VMS regions, but only as combined in a global VMS.

\section{Variational Autoencoders}
\label{VAE}

Autoencoders (AE) attempt to learn efficient latent-space encodings of the input data that would allow its reconstruction from such an encoding.  A variational autoencoder (VAE) \cite{kingma_auto-encoding_2013} is an extension of the AE, which has the training aim to maximise the lower bound of the marginal log-likelihood of the data following encoding and reconstruction. This means minimising the KL divergence between the posterior and {\em a priori} data distributions during the training. Rather than just learning a compressed encoding of the data, a VAE learns a probability distribution that is an approximation of the true probability distribution of the underlying data. This allows a VAE to be used as a generative model based on sampling in the latent space. 

VAEs are made up of two components - an \textit{encoder} which converts input data $x$ into a latent space representation $z$, and a \textit{decoder} that converts a latent space variable $z$ back into data $x'$ akin to the input $x$. CNNs are used for implementing both the encoder and the decoder. The encoder is defined as a probabilistic machine $q_{\theta}(z|x)$ that extracts a specific latent space code $z$ where $\theta$ represents the parameters of the encoder' network. Meanwhile, the decoder maps the information in a probabilistic sense defined by $p_{\phi}(x | z)$ in the opposite way from the code $z$ back to the data space $x$, where $\phi$ defined the parameters of the decoder network. The encoder and decoder are related through the loss function which consists of two components:
\begin{equation}
L(\theta,\phi) = -E_{z \sim q_{\theta}(z|x)}[\log p_{\phi}(x|z)] + KL(q_{\theta}(z|x)||p(z))
\label{eq1}
\end{equation}
where $KL(\cdot)$ represents the Kullback-Liebler divergence between the {\em a priori} distribution of the latent space $q_{\theta}(z|x_{i})$ and its estimated distribution $p(z)$.
The first term from equation (\ref{eq1}) represents the reconstruction loss and the second term regularises the learnt distribution. The latter term helps the VAE to learn to group conceptually similar data in the same regions of the latent space.

\vspace*{-0.4cm}
\section{Generating Visual Memory Schemas using VAEs}
\label{Sec4}
\vspace*{-0.1cm}

\begin{figure}[h]
\centering
\includegraphics[width=0.8\textwidth]{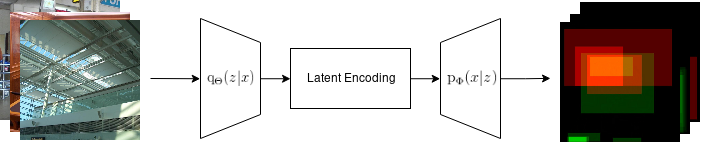}
\caption{Predicting VAEs in images using an autoencoder.}
\label{Fig2}
\vspace*{-0.3cm}
\end{figure}

The aim of this research study consists in developing a generative method for Visual Memory Schemas (VMS), for a given input image. In our approach we would aim to generate both true and false VMSs, simultaneously. This is defined as an image-to-image translation problem by making use of an VAE consisting of two CNNs, with the first one, the encoder designed to learn a mapping from an image to a latent code, while the decoder to learn the mapping from that latent code to a VMS. Previous work \cite{lukavsky_visual_2017,garcia-gasulla_behavior_2017} has shown that CNNs work well at extracting high-level image features that also allow for the prediction of memorability \cite{squalli-houssaini_deep_nodate}. CNNs such as VGG-16 network have also been shown to be capable of learning to reconstruct VMS maps at some degree for certain image categories \cite{akagunduz_defining_2019}. We propose using VAE models which have good ability to learn data classification in the latent space, as exemplified in Fig.~\ref{Fig2}. This model would allow a good separation of the false and positive VMS encoding spaces and then for the generation of dual channel VMS maps for generic scene input images corresponding to true and false VMS structures in which given random memorable images produce latent codes similar to those indicated experimentally by humans in memorable images. Moreover, the learned latent space modelled by VAEs can be easily inspected in order to find relations between the memorability and false memorability of images.

For the training we use a pre-trained VGG network architecture \cite{simonyan_very_2014} for the encoder after truncating the network before the classification step and using only the convolutional layers. The final output of the VGG architecture will be connected to a dense layer in order to compress the representation further, followed by the latent encoding. In CNNs the deep features that would emerge capture structures of the objects in the scene \cite{zhou_object_2014} and semantic structures \cite{gonzalez-garcia_semantic_2018} present in the input image.

The decoder would usually be simpler than the encoder. Whereas the input of the encoder consists of real world scenes, the output of the decoder is a VMS map, which consists of only two channels representing the spatial density of how likely a given image region is to cause that image to be remembered. There is no benefit in using a very deep architecture for the decoder, as we do not need to recreate any meaningful semantic features in the output. Additionally, a simpler architecture keeps the number of trainable parameters low, which is important when considering the low amount of available training data.

The loss function for this model is similar to the standard VAE loss function from (\ref{eq1}), with the exception that in the reconstruction term, instead of reconstructing the \textit{original} image data, aims to reconstruct associated information, such as VMSs. If $X$ is the set of scene images and $Y$ the set of associated VMS maps, with $x \in X$ and $y \in Y$ representing corresponding images drawn from these sets, our loss function is:
\begin{equation}
L = -E_{z \sim q_{\phi}(z|x)}[\log p_{\Psi}(y|z)] + KL(q_{\theta}(z|x)||p(z))
\label {eq2}
\end{equation}
where $\Psi$ represents the parameters associated with the VAE reconstructing the VMSs data $y$ at the end of the encoder.
We additionally investigate replacing the reconstruction term with the l1-norm as in \cite{akagunduz_defining_2019}.

\vspace*{-0.3cm}
\section{Experimental results}
\label{Exp}
\vspace*{-0.1cm}

\subsection{Experimental Setup}
\vspace*{-0.1cm}

For the encoder we use a pretrained VGG-16 network to extract a 
$7 \times  7 \times 512$ representation of an image, then compress this further using an $n$ dimensional dense layer, which leads to a latent space with a dimension of $m$. All parameters of the VGG network are frozen, by considering learning rates set to 0 during training, to avoid damaging the deep features while training on a small dataset such as ours. We employ data augmentation for training due to the small size of the training set. Data augmentation involves various realistic image manipulations, such as for example shifting the image either horizontally or vertically by 0.1 of the total image width, zooming the image, and horizontal flipping, for increasing the training data.

\begin{figure}[H]
\centering
\includegraphics[width=0.8\textwidth]{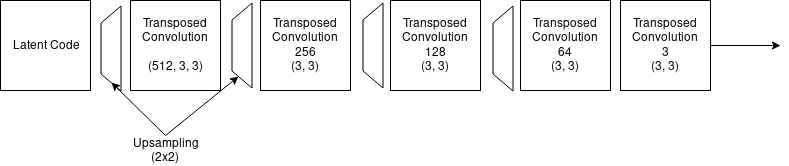}
\vspace*{-0.2cm}
\caption{Structure of the Decoder.}
\label{fig3}
\vspace*{-0.4cm}
\end{figure}

The decoder consists of a five layer upsampling network, shown in Fig.~\ref{fig3}, that implements transposed convolutions in order to convert the $m$-dimensional latent variable space back into an image. We apply batch normalisation after every convolution and employ l2 kernel regularisation \cite{l2reg}, l = 0.02, and a learning rate of 0.0001. We use a batch size of 32 and train the network for 250 epochs with 20 steps per epoch. In the experiments we evaluate three different architectures considering: 1) n = 64 and m=8; 2)  n = 64 and m=8
with an $l1$ reconstruction loss; 3) n = 128 and m=32.
The input and output of the entire architecture is a $224 \times 224$ image. The model is implemented in Keras\footnote{https://keras.io}.

\subsection{Datasets}
\vspace*{-0.1cm}

We evaluate the network over three datasets:
\begin{enumerate}
\item \textbf{VISCHEMA}. The dataset\footnote{https://www.cs.york.ac.uk/vischema/} used in \cite{akagunduz_defining_2019}. Consists of 800 images and 800 Visual Memory Schemas taken experimentally on a group of 100 people who were asked to indicate whether they remember certain images and if yes, to indicate what image regions made them remember them. This dataset is divided into a 640 image training set and a 160 image test set.
\item \textbf{VISCHEMA2}. A new set of scene images extracted from the FIGRIM dataset, and divided into the same hierarchical structure as the original VISCHEMA dataset. No ground truth visual memory schemas are available yet for this dataset, but because the categories and semantic content of the images are highly similar with the original dataset, VISCHEMA2 is useful for evaluation purposes.
\item \textbf{LaMem}. A dataset of 60,000 images, of a wide variety, with corresponding ground truth memorability scores \cite{khosla_understanding_2015}.
\end{enumerate}

\vspace*{-0.3cm}
\subsection{VMS reconstruction}
\vspace*{-0.1cm}

We evaluate reconstruction results of the original VISCHEMA dataset using both standard mean squared error (MSE) over all test images and the two dimensional Pearson product-moment correlation coefficients $\rho^{2D}$. We average the results on all true VMSs, and false VMSs, separately. True VMSs represent the VMS map regions indicated by participants in the experiments that represent what made them remember that image, while false VMSs represent regions from images, falsely indicated by people that made them remember those images. Actually those images have not been shown to them before. We obtain this metrics for all visual schemas and then evaluate the relation between this metric and the more standard `memorability score' provided in the LaMem dataset \cite{fajtl_amnet:_2018}. The relationship between visual memory schemas and computational saliency is also explored. Computational saliency maps for the VISCHEMA datasets are generated via the Graph Based Visual Saliency (GBVS) algorithm \cite{harel_graph-based_nodate}. 

Finally, we employ a state-of-the-art memorability prediction network and evaluate the relation between the VISCHEMA datasets memorability scores of the predicted VMS and the VMSs corresponding to the choices made by people, for both datasets, VISCHEMA and the VISCHEMA2. For all evaluations of our memorability metrics and standard memorability scores we follow prior work  from \cite{isola_what_2011}, \cite{khosla_understanding_2015} and use Spearmans rank correlation.

\begin{table}[h]
\centering
\begin{tabular}{|c|l|l|l|}
\hline
\multicolumn{1}{|l|}{Latent Space} & \multicolumn{1}{c|}{VMS}   & \multicolumn{1}{c|}{$\rho^{2D}$}  & \multicolumn{1}{c|}{MSE}    \\ 
\multicolumn{1}{|l|}{Dimension (m)}                  &       &       &  \\\hline
\multirow{3}{*}{32}        & True  & 0.545 & 92.54 \\ \cline{2-4} 
                              & False & 0.369 & 70.526 \\ \cline{2-4} 
                              & All   & \textbf{0.57} & 85.379   \\ \hline
\multirow{3}{*}{8}         & True  & 0.513 & 90.812 \\ \cline{2-4} 
                              & False & 0.333 & 64.228 \\ \cline{2-4} 
                              & All   & 0.53  & 83.472 \\ \hline
\multirow{3}{*}{8 and L1 norm in (\ref{eq2})}      & True  & 0.543 & 72.348 \\ \cline{2-4} 
                              & False & 0.168 & 25.131 \\ \cline{2-4} 
                              & All   & 0.559 & 72.052 \\ \hline
\end{tabular}
\vspace{0.4cm}
\caption{Reconstruction accuracy for three deep learning architectures.}
\label{tab:reconacc}
\vspace{-0.3cm}
\end{table}

Table~\ref{tab:reconacc} shows the reconstruction results in terms of both MSE and Spearmans rank correlation, $\rho^{2D}$. The network with an m=8 dimensional latent space and an l1-norm component to its loss function has the overall best MSE, while the network with the overall best Pearsons correlation with the ground truth is the network with a m=32 dimensional latent space. Our overall $\rho^{2D}$ results are slightly worse than those presented in \cite{akagunduz_defining_2019}, though it should be noted that we generate both the true and false maps simultaneously. This allows us to investigate how well the individual true and false VMS are reconstructed.
In general, false VMS maps are more difficult to accurately reconstruct than true VMS maps. This is likely due to the overall lower consistency between human observers for false VMS maps. While what is memorable tends to be well agreed on among people, what causes false remembering of an image is more varied, and this effect crosses over to generative models. Interestingly, we find that a higher dimensional latent space has the best effect on reconstruction accuracy, rather than the use of an l1-norm in the loss term. This is due to the effect of the second term in the loss function from equation (\ref{eq2}) and indicates that higher dimensional spaces are better at capturing `memorability'. 
For the rest of this section we evaluate the results of the network with a $m=32$ dimensional latent space, given that this architecture performs the best as measured by the $\rho^{2D}$ metric.

\begin{figure}[H]
\vspace{-0.2cm}
\centering
\includegraphics[width=0.7\textwidth]{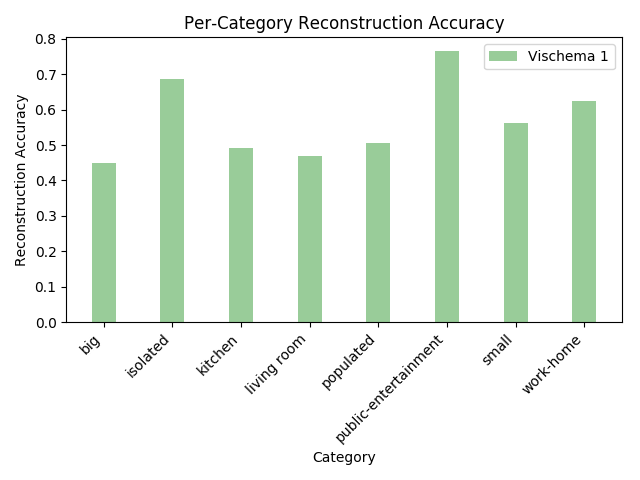}
\vspace{-0.2cm}
\caption{Reconstruction accuracy for various image categories.}
\label{fig:viscrecon}
\vspace{-0.4cm}
\end{figure}

Figure~\ref{fig:viscrecon} shows the reconstruction accuracy measured by $\rho^{2D}$ for each category in the VISCHEMA dataset, over the 160 image test set. We find that the best performing category is that of Public Entertainment, with a correlation of $0.766$, which is better that the results from \cite{akagunduz_defining_2019} which found that the Work-Home image category had the best performance with a correlation of $0.677$. A comparison with prior work is shown in Table \ref{tab:tabular_compare}.

\begin{table}[H]
\centering
\resizebox{\textwidth}{!}{%
\begin{tabular}{|l|l|l|l|l|l|}
\hline
Work & Best Category & $\rho^{2D}$ & Worst Category & $\rho^{2D}$ & Overall $\rho^{2D}$ \\ \hline
Previous Method & Work/Home & 0.677 & Living Room & 0.506 & 0.588 \\ \hline
Our Method & Public Entertainment & 0.766 & Big & 0.449 & 0.57 \\ \hline
\end{tabular}%
}
\vspace*{0.1cm}
\caption{Comparison with Prior Work}
\label{tab:tabular_compare}
\end{table}
\vspace*{0.2cm}

The worst performing category for VMS reconstruction is the "Big" which contains images of airport terminals with a correlation of $0.449$. In general, we find that categories that have high overall memorability tend to reconstruct better than the categories with low overall memorability. Differences from prior work may also be due to generating higher resolution images, which captures more detail in some categories yet causes more divergence in categories with less available memorability information. We found that the correlation between predicted VMS maps and saliency maps, provided by GBVS algorithm  \cite{harel_graph-based_nodate}, to be $0.69$ which agrees with other results on the relationship between memorability and saliency  \cite{object_mem,akagunduz_defining_2019}. GBVS is a well used saliency measure, but VMS maps offer more than saliency alone. When averaging on all image categories and comparing with saliency, we found that false VMS maps have a correlation of $0.625$ while true VMS maps have a correlation of $0.704$.

\vspace*{-0.2cm}
\subsection{Memorability results}
\vspace*{-0.2cm}

\begin{figure}[h]
\centering
\begin{tabular}{cc}
\includegraphics[width=0.4\textwidth]{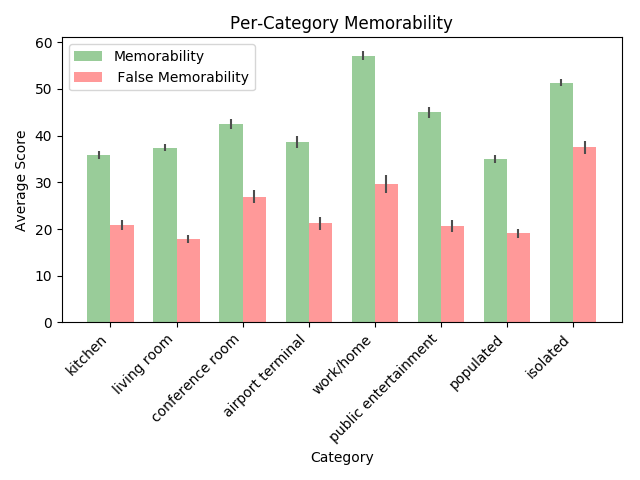}  &
\includegraphics[width=0.4\textwidth]{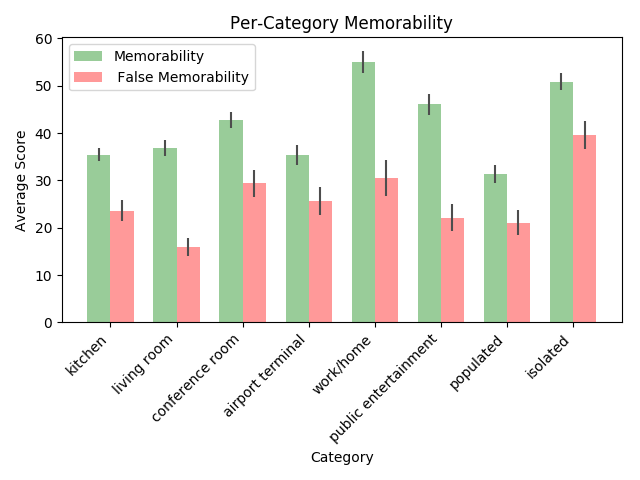}\\
(a) VISCHEMA2  & (b) VISCHEMA \\
\end{tabular}\\
\vspace*{0.2cm}
\caption{Comparison of the memorability results for a set of image categories between the VISCHEMA2 and VISCHEMA datasets.}
\label{fig:visc}
\vspace*{-0.3cm}
\end{figure}

\begin{figure}[h]
\vspace*{-0.3cm}
\centering
\begin{tabular}{cc}
\includegraphics[width=0.49\textwidth]{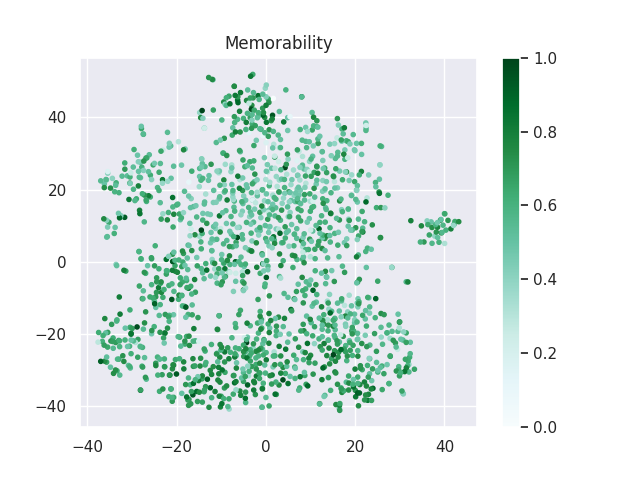} &
\includegraphics[width=0.49\textwidth]{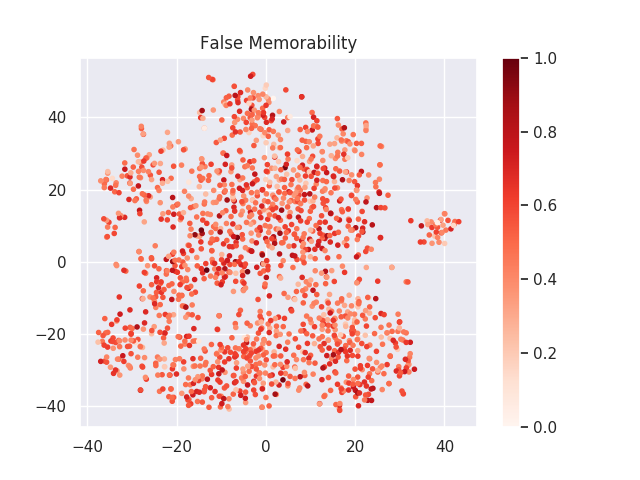} \\
(a) Memorability  & (b) False memorability \\
\vspace*{0.07cm}
\end{tabular}
\caption{VISCHEMA2 Latent Space Embedding. Green represents memorability and red represents false memorability, normalised between 0 and 1.}
\label{fig:viscembed}
\end{figure}

\begin{figure}[h]
\centering
\includegraphics[width=0.7\textwidth]{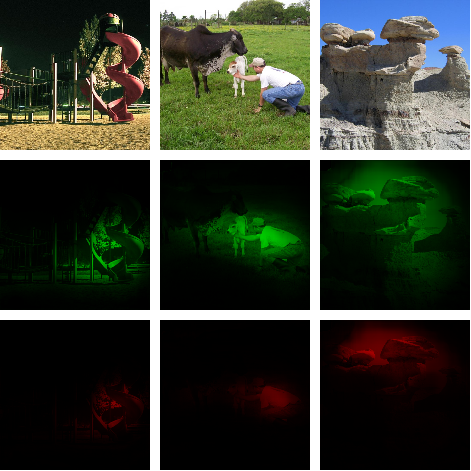}
\caption{Set of three images from VISCHEMA2 dataset and their predicted true VMS and false VMS on second and third lines.}
\label{fig:viscpred}
\vspace*{-0.5cm}
\end{figure}

We generate 800 predicted VMS maps for the 800 images in the VISCHEMA2 dataset and find that the distribution of memorability and false memorability agrees with that of the original ground truth dataset, according to the results from Fig.~\ref{fig:visc} with Spearmans ranks of $0.929$ and $1.0$, respectively for $p < 0.01$. This is due to the similarity of the datasets, but it also shows that the proposed model has successfully learned to generate VMSs that agree on a category-wide scale despite being trained with no category labels. Additionally, we find that in general the higher the memorability of an image, the higher its own false memorability, as we can observe from the similarity of the clusters of the latent space embeddings of the Memorability and those corresponding to False Memorability, shown in Fig.~\ref{fig:viscembed}a and \ref{fig:viscembed}b, respectively. Images that tend to be highly memorable also tend to be highly falsely memorable. In Fig.~\ref{fig:viscpred},  three images from VISCHEMA2 are shown on first line and their corresponding true and false VMSs are shown on second and third line, respectively.

\vspace*{-0.5cm}

\subsection{VMS Maps and Memorability Scores}

Predicted memorability scores for both VISCHEMA 1 and 2 datasets were obtained by employing the AMNet network \cite{fajtl_amnet:_2018}. These scores were then compared to the memorability metric used for evaluating visual schemas. No significant relationship was found between the per-category memorability metrics and the predicted category memorability scores aside from VISCHEMA2's "Populated" category which had a Spearmans rank correlation with the AMNet scores of $0.203$ with $p < 0.01$. It appears that VMSs, even predicted schemas, do not directly relate to predicted memorability scores for the same images, and that unlike our VMS prediction model, predicted memorability scores may not take fully into account what humans find memorable. It has been shown that deep neural networks take the simplest approach possible to solving a problem \cite{neuralnetspuzzling}, and it is possible that memorability prediction models are working on factors that do not necessarily align directly with memorability if some other learned metric provides a `good enough' approximation. This could explain why predicted scores do not align with VMS maps.

We also examine the relationship between the ground truth memorability scores and our metric by predicting VMSs for a 10,000 image subset of the LaMem dataset, used in \cite{fajtl_amnet:_2018}, and estimating only the true memorability score for them. We then use the Spearmans rank to compare the ground-truth score and our metric. We find a rank correlation of $0.147$ with $p < 0.01$, indicating that VMS maps and experimentally-based memorability scores are weakly, but significantly, related.

\vspace*{-0.3cm}
\section{Conclusion}
\label{Con}
\vspace*{-0.2cm}

In this research study we have constructed and evaluated a VAE model capable of predicting Visual Memory Schemas for a given input image. The VAE model is used for generating both true and false VMS maps simultaneously at over ten times the resolution of previous approaches. Moreoever, we find a very close correlation between the ground truth per-category metrics and the predicted per-category metrics, and finally show that current state-of-the-art memorability prediction does not appear to correlate with ground truth or predicted VMS metrics, and that these metrics do have a significant, but weak, positive correlation with ground truth memorability scores from the LaMem dataset. This indicates that VMSs can provide additional information about image memorability which is not traditionally captured by other memorability prediction methods.

\vspace*{-0.4cm}
\section{Acknowledgements}
\label{Sec5}
\vspace*{-0.2cm}

The first author would like to acknowledge the support from the Doctoral Training Grant provided by the EPSRC.

This project was undertaken on the Viking Cluster, which is a high performance compute facility provided by the University of York.  We are grateful for computational support from the University of York High Performance Computing service,Viking and the Research Computing team.

\vspace*{-0.5cm}
\bibliography{vms_vae_paper}

\end{document}